\pdfoutput=1

\documentclass[10pt,twocolumn,letterpaper]{article}

\usepackage[pagenumbers]{cvpr} 

\usepackage{graphicx}
\usepackage{amsmath}
\usepackage{amssymb}
\usepackage{booktabs}
\usepackage{amsmath}
\usepackage{amssymb}
\usepackage{makecell}
\usepackage{caption}
\usepackage{times}
\usepackage[pagebackref,breaklinks,colorlinks]{hyperref}

\usepackage[capitalize]{cleveref}
\crefname{section}{Sec.}{Secs.}
\Crefname{section}{Section}{Sections}
\Crefname{table}{Table}{Tables}
\crefname{table}{Tab.}{Tabs.}

\def\cvprPaperID{18} 
\def\confName{CVPR}
\def\confYear{2022}

\begin{document}

\title{StyLandGAN: A StyleGAN based Landscape Image Synthesis using Depth-map}
\author{Gunhee Lee \and Jonghwa Yim \and Chanran Kim \and Minjae Kim \and 
Vision AI Lab, AI Center, NCSOFT \\
{\tt\small \{victorleee, jonghwayim, chanrankim, minjaekim\}@ncsoft.com}
}

\twocolumn[{%
\renewcommand\twocolumn[1][]{#1}%
\maketitle
\begin{center}
    \centering
    \captionsetup{type=figure}
    \includegraphics[width=2\columnwidth]{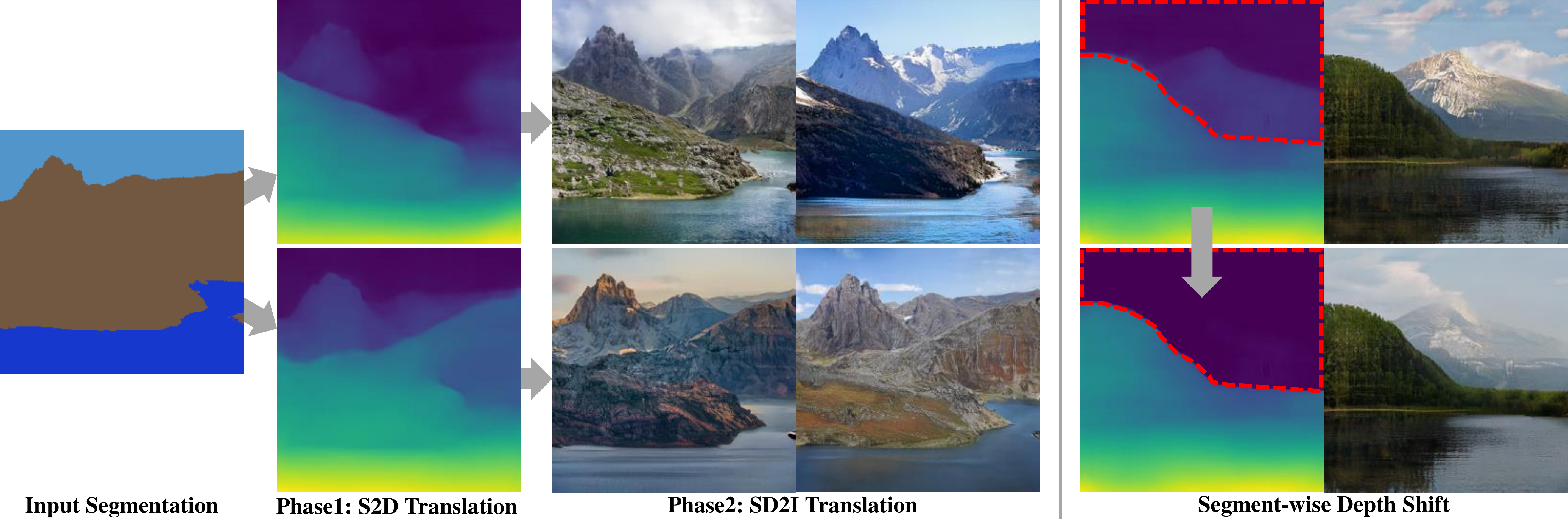}
    \captionof{figure}{Our proposed StyLandGAN framework can synthesize diverse structure and style landscape images from a single segmentation map. \textbf{(Left)} Our `2-phase inference' firstly generates diverse depth maps from a single segmentation map (Phase1:S2D), and then generates landscape images from both segmentation and depth map inputs (Phase2:SD2I). \textbf{(Right)} We also show that shifting the local region of the depth map (Here, the sky and mountain areas) can achieve the desired result.}
    \label{fig:teaser}
\end{center}%
}]

\begin{abstract}
Despite recent success in conditional image synthesis, prevalent input conditions such as semantics and edges are not clear enough to express `Linear (Ridges)' and `Planar (Scale)' representations. To address this problem, we propose a novel framework StyLandGAN, which synthesizes desired landscape images using a depth map which has higher expressive power. Our StyleLandGAN is extended from the unconditional generation model to accept input conditions. We also propose a '2-phase inference' pipeline which generates diverse depth maps and shifts local parts so that it can easily reflect user's intend. As a comparison, we modified the existing semantic image synthesis models to accept a depth map as well. Experimental results show that our method is superior to existing methods in quality, diversity, and depth-accuracy.
\end{abstract}

\section{Introduction}
Visual contents creation tools have been developed to allow users to embody their own abstract ideas into concrete images. In particular, tools for landscape creation are also spotlighted in the field of environment concept art, which shows a glimpse of world in computer games~\cite{park2019SPADE}. These tools are developed mainly in the name of `conditional image synthesis' based on GANs~\cite{NIPS2014_5ca3e9b1}. 
 
\begin{figure}[t]
\begin{center}
\includegraphics[scale=0.5]{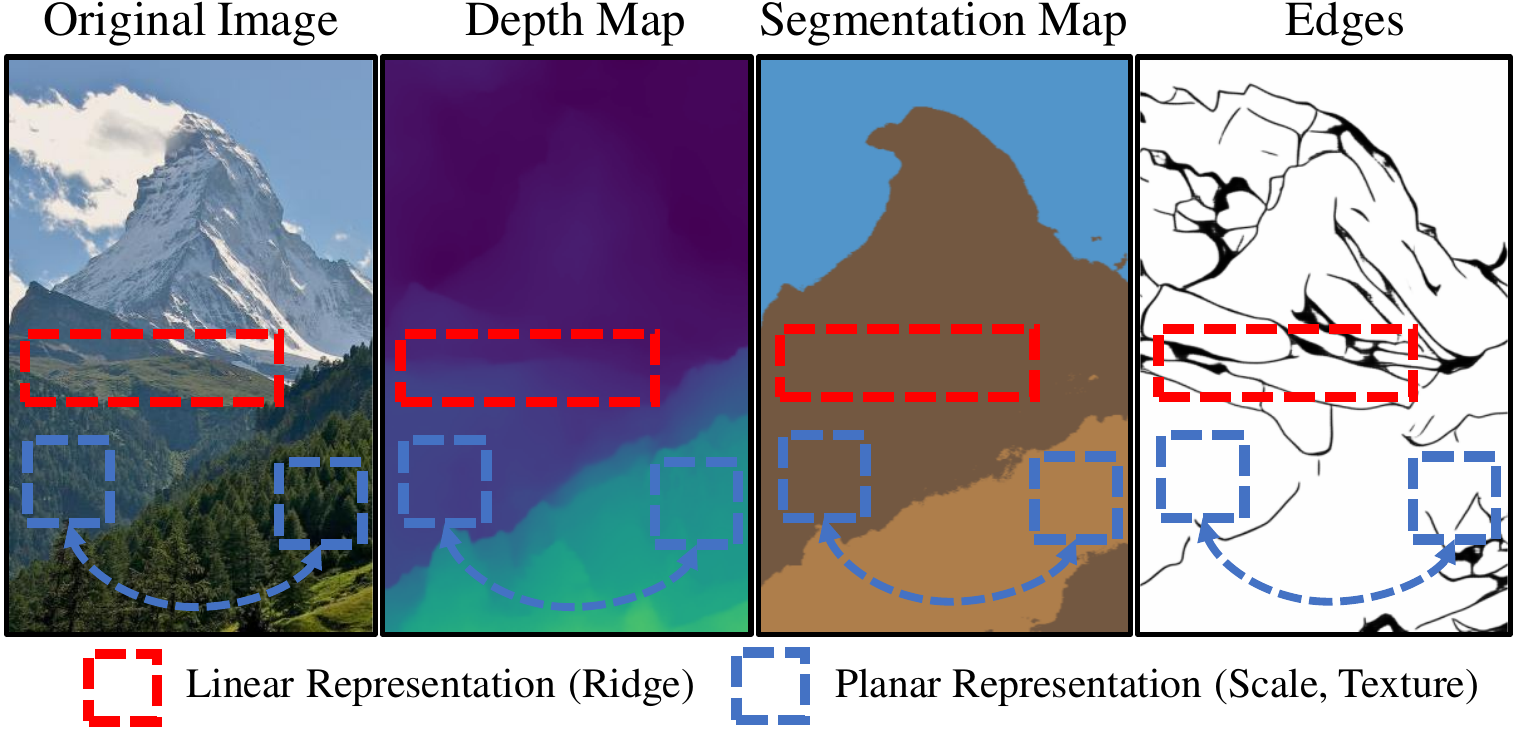}
\end{center}
   \caption{Comparison between various input modalities. A depth map can express both ridge (red) and scale (blue) information. Segmentation is hard to handle ridge, and edge is difficult to express scale information.}
\label{fig:intro_segmap}
\end{figure}

Although previous conditional models can synthesize visually pleasing landscape images either from segmentation map or edge scribbles (Sec.~\ref{related_work0}), these conditional inputs are not sufficient to convey user's intention due to low expressive power. Comparison between conditional inputs is shown in Figure \ref{fig:intro_segmap}. Although most of the previous works use \textbf{segmentation map} to control the contents of the image, it does not contain both linear (Ridge) and planar (Scale, Texture) representations. There are also works exploiting \textbf{edge scribbles} to control images. Even if humans can perceive depth from edge drawings ~\cite{DBLP:journals/corr/abs-2002-06260}, the edge itself lacks planar representation which leads to depth ambiguity. Therefore, synthesizing landscape images only from those inputs leaves depth representation unclear. Unlike these input modalities, we observe that \textbf{depth map} can express both linear and planar representations. Furthermore, with recent improvements in monocular depth map estimation~\cite{Miangoleh2021Boosting,GargDualPixelsICCV2019,monodepth2,Ranftl2020}, it becomes much easier to acquire a pseudo depth map even from a single image. Thus, one can build an image-depth pair dataset without any ground truth depth maps.


\begin{figure*}
\begin{center}
\includegraphics[width=2\columnwidth]{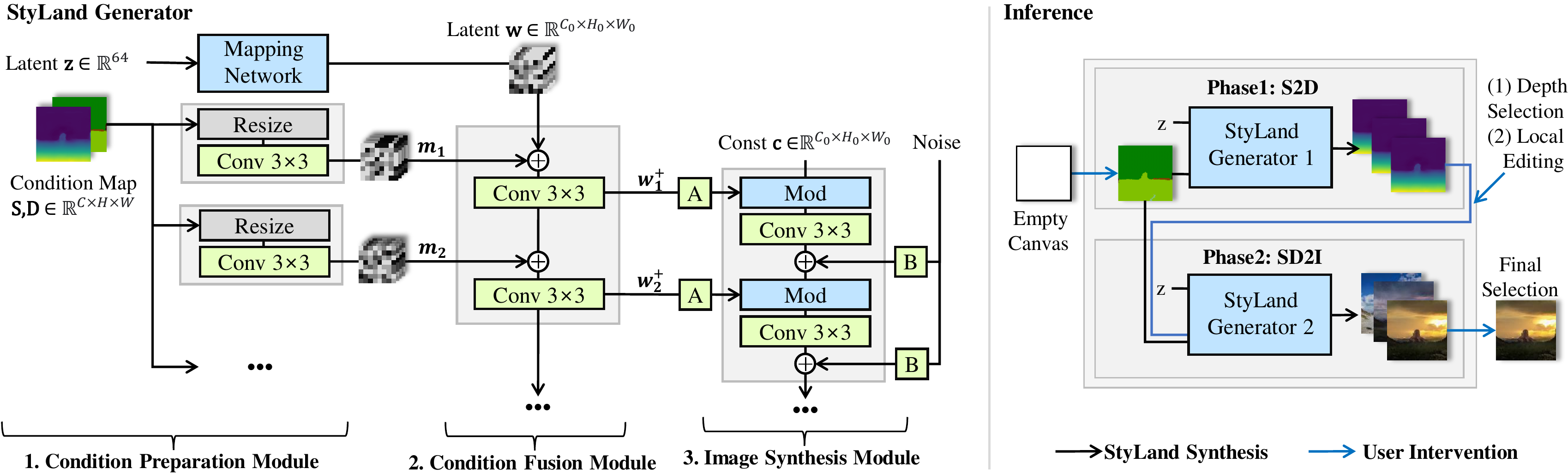}
\end{center}
   \caption{(Left) Our StyLand Generator composed of 3 different modules: condition preparation module, condition fusion module, and image synthesis module. (Right) Our `2-phase inference' first generates depth of given segmentation map then generates final landscape output.}
\label{fig:overall}
\end{figure*}

 In this paper, we present the StyLandGAN framework that uses multiple input conditions including segmentation and depth map to convey user intentions. Our framework extends the previous StyleMapGAN~\cite{kim2021stylemapgan} as a conditional-GAN framework. In order to synthesize landscape images, we first suggest a segmentation and depth-to-image translation model (SD2I). Moreover, we propose a segmentation-to-depth translation model (S2D) which provides diverse global depth suggestions for users to acquire the desired depth map more easily. In addition, we design a segment-wise depth shift technique to enable local region editing. Combining all of the above, we provide a `2-phase inference' pipeline that first generates structure information using S2D translation, edits local parts if needed, and finally generates landscape images using SD2I translation. Experimental results show that StyLandGAN outperforms previous works with an identical depth map input setting.

\section{Related Work}

\label{related_work0}
 \textbf{Image Synthesis Approach.} Image synthesis has been developed using two different approaches. The first approach is the \emph{create-from-simple} approach. This method generates a complex image from a relatively simple condition input. This scheme can have various input modalities, from semantic segmentation map~\cite{esser2020taming,park2019SPADE,pix2pix2017,liu2019learning,schnfeld2021you}, edges~\cite{pix2pix2017,Chen_2018_CVPR}, or even from text prompts~\cite{pmlr-v139-ramesh21a,pmlr-v48-reed16,ming2020DFGAN}. There are also methods using various conditional inputs ~\cite{Nederhood_2021_ICCV,esser2020taming,huang2021multimodal}. Secondly, with the \emph{choose-from-many} approach, multiple images may be created from a single input.  In the line of research on image-to-image translation (I2I), many approaches have been developed not only for single-modal translation~\cite{pix2pix2017,wang2018pix2pixHD,park2019SPADE,Zhu_2020_CVPR,zhang2020cross,Kim2020U-GAT-IT:} but also for multi-modal translation~\cite{huang2018munit,DRIT,DRIT_plus,zhu2017toward,choi2020starganv2,saito2020coco}. Our framework is designed to use both approaches, which starts with simple condition input, then outputs relatively complex multiple outputs.

 \textbf{Representation Disentanglement.} Previous I2I translation research has attempted to separate `content information' and `style information' in the image ~\cite{huang2018munit,DRIT,DRIT_plus,choi2020starganv2,saito2020coco,park2020swapping}. However, this separation is an ill-defined and ill-posed problem. As suggested in ~\cite{Nederhood_2021_ICCV}, we also define the conditioning information as `content'. This approach leads to define `style' as information that contents do not capture. To be more specific in our segmentation and depth-to-image translation (SD2I) model, it can be said that depth and segmentation maps are mainly responsible for content information and the rest are style information.
 
\section{StyLandGAN}
Our goal is to obtain a model capable of generating diverse images with a given conditions. Here, we describe our proposed model that exploits two different modalities, i.e., semantic segmentation and depth map. We propose StyLandGAN which adopts a \emph{condition latent code}, an intermediate 2D spatial latent code representing segmentation and depth maps. We denote a single image as $\mathcal{I}$, and the corresponding segmentation map and depth map as $\mathcal{S}$, and $\mathcal{D}$, respectively. 

\subsection{StyLand Generator}

  Our generator consists of three components as shown in Figure~\ref{fig:overall}: (1) Condition preparation module, (2) Condition fusion module, and (3) Image synthesis module. These three modules are composed separately, but sequentially the previous result is given to the next module to synthesize the final image.
  
  \textbf{Condition preparation module} prepares random latent code and condition latent code separately. Thus, this module consists of mapping network for random latent code and condition blocks for condition latent codes. As in ~\cite{kim2021stylemapgan}, our mapping network produces random latent code $w \in \mathbb{R}^{{C}_0\times {H}_0\times {W}_0}$ with initial spatial size. Each condition blocks firstly resizes input condition $\mathcal{S}, \mathcal{D}\in \mathbb{R}^{{C}\times H\times W}$ into $\mathcal{S}_i, \mathcal{D}_i\in \mathbb{R}^{C_i\times H_i\times W_i}$ for each $i$-th layer. Then, these resized maps are concatenated in channel-wise direction, and output condition latent code $m_i \in \mathbb{R}^{C_i\times H_i\times W_i}$ by convolution operation.

\textbf{Condition fusion module} merges random latent code $w$ and condition latent code $m_1$ of the first layer by element-wise addition operation. Then, it outputs the intermediate latent code, $w^+_1$ by convolution operation for next layer. All the intermediate latent code $w^+_i\in \mathbb{R}^{C_i\times H_i\times W_i}$is passed to the $i$-th layer of image synthesis module, and also fused with the ${(i+1)}$-th condition latent code, $m_{i+1}$. This operation continues till it reaches the size of original image $\mathcal{I}\in \mathbb{R}^{{C}\times H\times W}$ to convey more detailed spatial information.
  
  Finally, \textbf{Image synthesis module} progressively synthesizes image with affine transformed latent code of each layer, $w^+_i$. We add per-pixel noise which is introduced in \cite{Karras2019stylegan2}, so that it can generate a diverse and delicate changes of landscape images.

  We follow the training scheme procedure in StyleMapGAN ~\cite{kim2021stylemapgan} training scheme which uses adversarial loss~\cite{NIPS2014_5ca3e9b1}, R1 regularization~\cite{Mescheder2018ICML}, Perceptual loss~\cite{zhang2018perceptual}, Domain-guided loss~\cite{zhu2020indomain}, and reconstruction loss. Like ~\cite{kim2021stylemapgan}, we also use the encoder and jointly train with generator and discriminator.

\subsection{Depth Map Acquisition}
As we have argued in Figure~\ref{fig:intro_segmap}, a depth map is beneficial for handling linear and planar representation. However, it  requires additional delicate effort to obtain the desired depth map. Thus, we suggest the global acquisition method `S2D translation' and the local editing method `Segment-wise depth shift' to acquire the desired depth map.

\textbf{Global: S2D translation.} We have argued a basic model that translates the segmentation map and depth map into the corresponding image, which is the SD2I translation model. Likewise, we also train the segmentation-to-depth (S2D) translation model, with an identical StyLand generator. Through this model, users can acquire diverse corresponding depth maps from single segmentation input.

\textbf{Local: Segment-wise depth shift.} Here, we suggest a simple method to edit the selected global depth map. We observed that the semantic label `sky' have smaller depth values than the semantic label `grass'. In other words, the local depth value can be shifted to a certain level only to the extent that the relative depth order of semantics does not change. Under this constraint, we show that shifting the depth of local parts can change the local view of a landscape. The representative example is depicted in Figure \ref{fig:teaser}.
 
\subsection{Two-phase inference}
As a final inference model, we suggest a `2-phase inference' as shown in Figure~\ref{fig:overall}. It consists of two StyLand generators, i.e., S2D and SD2I.  The depth maps are synthesized from  user drawn semantic segmentation map by the first generator (Phase1), and both segmentation and selected depth are passed to the second generator to produce final landscape image (Phase2). Note that one can now easily get the desired depth map by selecting a candidate from diverse depth suggestions and by editing it locally. Since depth map mainly determines the structure of a landscape, the proposed inference model can be divided into `structure selection' and `style selection', respectively.

\begin{figure*}
\begin{center}
\includegraphics[width=2\columnwidth]{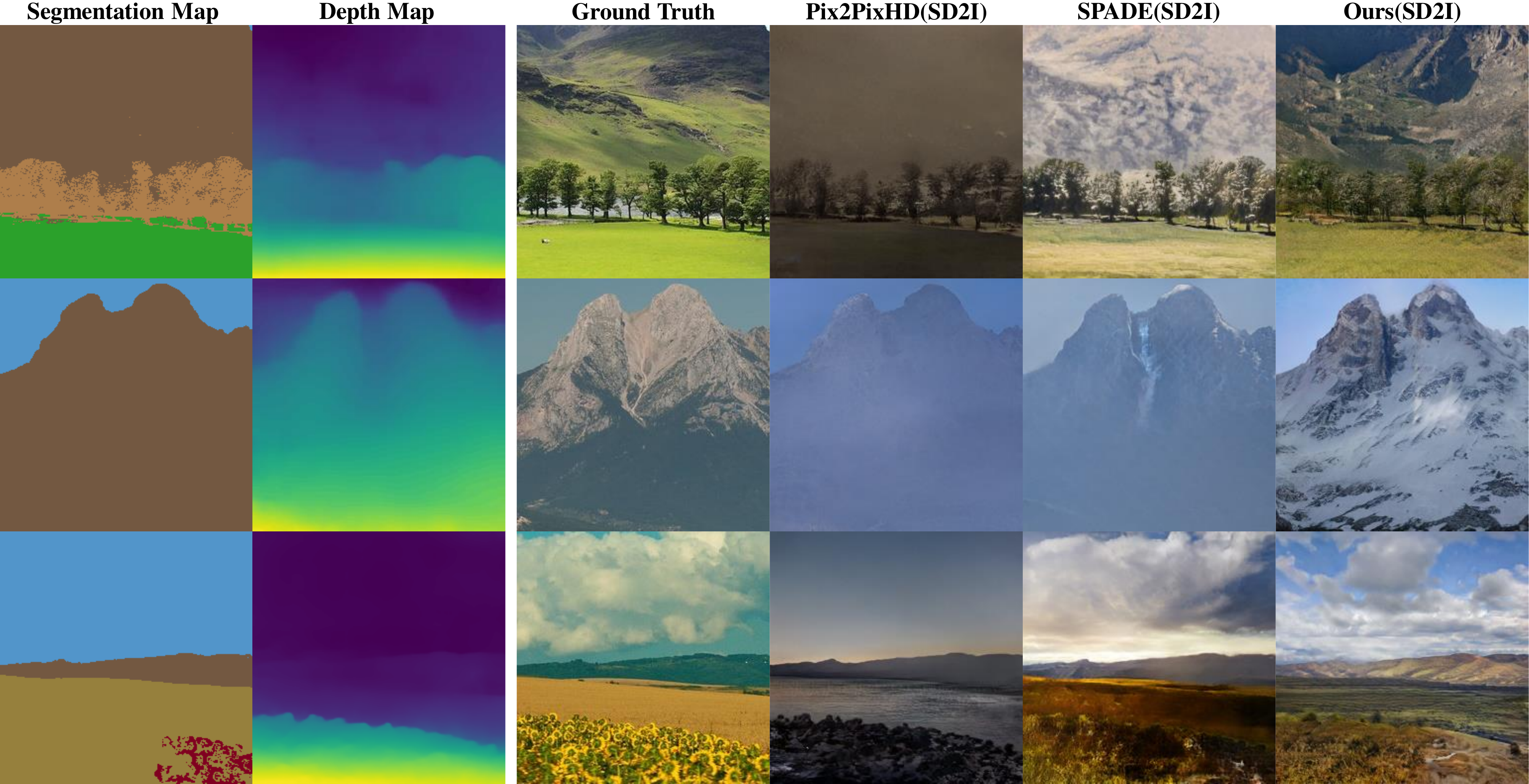}
\end{center}
   \caption{Qualitative comparison of SD2I translation model on Flickr-Landscape. Our model produces images with higher visual quality. More results are shown in supplementary materials.}
\label{fig:result}
\end{figure*}

\section{Experiments and Results}
\textbf{Implementation details.} Our StyLandGAN uses an input image size of $256\times256$ for all experiments. The shape of the random latent code is set to $w \in \mathbb{R}^{64\times8\times8}$. We use the ADAM optimizer with $(\beta_1,\beta_2)=(0.0,0.99)$, with a learning rate of 0.002, and a batch size of 8.

\textbf{Dataset.} Like all other previous methods ~\cite{park2019SPADE,esser2020taming,park2020swapping,Logacheva_2020_ECCV}, we also scrape landscape images from \href{http://www.flickr.com}{flickr.com} to construct a dataset named Flickr-Landscape. We randomly collected 48K images as a training set and 1K images as a test set. To build a segmentation-depth pair sets, we use the monocular depth estimation module ~\cite{Miangoleh2021Boosting}, and semantic segmentation estimation module \cite{zhang2020resnest}.
\begin{table}
\begin{center}
\begin{tabular}{l c c c}
\Xhline{2\arrayrulewidth}
\textbf{S2I Model} & FID($\downarrow$) & LPIPS($\uparrow$) & RMSE($\cdot$)\\
\hline
Pix2PixHD~\cite{wang2018pix2pixHD} & 56.08& N/A & 9.40\\
SPADE~\cite{park2019SPADE} & 47.34 & 0.39 & 9.43\\
CC-FPSE~\cite{liu2019learning} &  48.43& 0.18 & 9.29\\
OASIS~\cite{schnfeld2021you} &  \textbf{31.92} & \underline{0.51} & 9.34\\
StyLandGAN (Ours) & \underline{34.50} & \textbf{0.56} & 9.36\\
\Xhline{2\arrayrulewidth}
 &  & &\\
\Xhline{2\arrayrulewidth}
\textbf{SD2I Model} & FID($\downarrow$) & LPIPS($\uparrow$) & RMSE($\downarrow$)\\
\hline
Pix2PixHD~\cite{wang2018pix2pixHD} & 48.22& N/A & \underline{8.95}\\
SPADE~\cite{park2019SPADE} & \underline{43.33} & \underline{0.23} & 9.10\\
StyLandGAN (Ours) & \textbf{33.60} & \textbf{0.45} & \textbf{8.80}\\
\Xhline{2\arrayrulewidth}
\end{tabular}
\end{center}
\caption{Quantitative results of S2I and SD2I (Phase2 of 2-phase inference) model on Flickr-Landscape. The best scores are highlighted in bold, and the second-best are underlined. The arrow $\downarrow$ (or $\uparrow$) indicates that a lower (or higher) score is better.}
\label{table:quant_results}
\end{table}

 \textbf{Baselines.} To compare the performance of the proposed method, we first select previous segmentation-to-image(S2I) translation models: Pix2PixHD~\cite{wang2018pix2pixHD}, SPADE~\cite{park2019SPADE}, CC-FPSE~\cite{liu2019learning}, and OASIS~\cite{schnfeld2021you}. Secondly, we extended Pix2Pix and SPADE for the SD2I translation task to compare our model fairly. The same concatenation approach for multi-modal inputs is applied to all methods.

 \textbf{Metrics.} As other I2I models, we adopt FID~\cite{fidmetric2017} to measure `image quality'. We evaluate the `image diversity' by calculating the mean LPIPS~\cite{zhang2018perceptual} distance between generated images. For each test image, we calculated the mean distance between 10 randomly generated images with fixed semantic and depth maps. We also measure `depth accuracy' by using RMSE~\cite{diode_dataset} between the depth map of the original image and the depth map of the synthesized image.
 
 \textbf{Results.} In Table \ref{table:quant_results}, we present the summarized comparison results on S2I and SD2I translation separately. Note that Pix2PixHD can generate only a single image, and depth accuracy results on the S2I models are attached for reference. For the S2I model, StyLandGAN shows comparable score to other models. For the SD2I model, the proposed method achieves the best score in terms of FID, LPIPS, and RMSE. More importantly, we can observe the effectiveness of depth map as an additional information by comparing results between S2I and SD2I model. The depth guidance restricts the amount of randomness (image diversity) and enforces to generate intended representation (depth accuracy). Thus, for the SD2I model, the overall image diversity and depth accuracy score decreased compared to the S2I model. As shown in Figure ~\ref{fig:result}, our model produces landscape images corresponding to input semantic map, and synthesizes more details in the virtue of plausible linear and planar representations from depth map.

\section{Conclusion}

 In this paper, we present a novel conditional landscape synthesis framework, StyLandGAN, with depth map which is capable of expressing ridge and scale representation. We show that our `2-phase inference' makes it possible to acquire diverse structure and style of landscape images in a row. Our framework exceeds previous I2I translation method in image quality, image diversity, and depth-accuracy.

{\small
\bibliographystyle{ieee_fullname}
\bibliography{egbib}

\begin{thebibliography}{10}\itemsep=-1pt

\bibitem{Chen_2018_CVPR}
Wengling Chen and James Hays.
\newblock Sketchygan: Towards diverse and realistic sketch to image synthesis.
\newblock In {\em The IEEE Conference on Computer Vision and Pattern
  Recognition (CVPR)}, June 2018.

\bibitem{choi2020starganv2}
Yunjey Choi, Youngjung Uh, Jaejun Yoo, and Jung-Woo Ha.
\newblock Stargan v2: Diverse image synthesis for multiple domains.
\newblock In {\em Proceedings of the IEEE Conference on Computer Vision and
  Pattern Recognition}, 2020.

\bibitem{esser2020taming}
Patrick Esser, Robin Rombach, and Björn Ommer.
\newblock Taming transformers for high-resolution image synthesis, 2020.

\bibitem{GargDualPixelsICCV2019}
Rahul Garg, Neal Wadhwa, Sameer Ansari, and Jonathan~T. Barron.
\newblock Learning single camera depth estimation using dual-pixels.
\newblock {\em ICCV}, 2019.

\bibitem{monodepth2}
Cl{\'{e}}ment Godard, Oisin {Mac Aodha}, Michael Firman, and Gabriel~J.
  Brostow.
\newblock Digging into self-supervised monocular depth prediction.
\newblock October 2019.

\bibitem{NIPS2014_5ca3e9b1}
Ian Goodfellow, Jean Pouget-Abadie, Mehdi Mirza, Bing Xu, David Warde-Farley,
  Sherjil Ozair, Aaron Courville, and Yoshua Bengio.
\newblock Generative adversarial nets.
\newblock In Z. Ghahramani, M. Welling, C. Cortes, N. Lawrence, and K.~Q.
  Weinberger, editors, {\em Advances in Neural Information Processing Systems},
  volume~27. Curran Associates, Inc., 2014.

\bibitem{DBLP:journals/corr/abs-2002-06260}
Aaron Hertzmann.
\newblock Why do line drawings work? {A} realism hypothesis.
\newblock {\em CoRR}, abs/2002.06260, 2020.

\bibitem{fidmetric2017}
Martin Heusel, Hubert Ramsauer, Thomas Unterthiner, Bernhard Nessler, , and
  Sepp Hochreiter.
\newblock Gans trained by a two time-scale update rule converge to a local nash
  equilibrium.
\newblock In {\em Advances in Neural Information Processing Systems 30: Annual
  Conference on Neural Information Processing Systems 201}, 2017.

\bibitem{huang2018munit}
Xun Huang, Ming-Yu Liu, Serge Belongie, and Jan Kautz.
\newblock Multimodal unsupervised image-to-image translation.
\newblock In {\em ECCV}, 2018.

\bibitem{huang2021multimodal}
Xun Huang, Arun Mallya, Ting-Chun Wang, and Ming-Yu Liu.
\newblock Multimodal conditional image synthesis with product-of-experts gans.
\newblock In {\em Proc. CVPR}, 2021.

\bibitem{pix2pix2017}
Phillip Isola, Jun-Yan Zhu, Tinghui Zhou, and Alexei~A Efros.
\newblock Image-to-image translation with conditional adversarial networks.
\newblock {\em CVPR}, 2017.

\bibitem{Karras2019stylegan2}
Tero Karras, Samuli Laine, Miika Aittala, Janne Hellsten, Jaakko Lehtinen, and
  Timo Aila.
\newblock Analyzing and improving the image quality of {StyleGAN}.
\newblock In {\em Proc. CVPR}, 2020.

\bibitem{kim2021stylemapgan}
Hyunsu Kim, Yunjey Choi, Junho Kim, Sungjoo Yoo, and Youngjung Uh.
\newblock Exploiting spatial dimensions of latent in gan for real-time image
  editing.
\newblock In {\em Proceedings of the IEEE Conference on Computer Vision and
  Pattern Recognition}, 2021.

\bibitem{Kim2020U-GAT-IT:}
Junho Kim, Minjae Kim, Hyeonwoo Kang, and Kwang~Hee Lee.
\newblock U-gat-it: Unsupervised generative attentional networks with adaptive
  layer-instance normalization for image-to-image translation.
\newblock In {\em International Conference on Learning Representations}, 2020.

\bibitem{DRIT}
Hsin-Ying Lee, Hung-Yu Tseng, Jia-Bin Huang, Maneesh~Kumar Singh, and
  Ming-Hsuan Yang.
\newblock Diverse image-to-image translation via disentangled representations.
\newblock In {\em European Conference on Computer Vision}, 2018.

\bibitem{DRIT_plus}
Hsin-Ying Lee, Hung-Yu Tseng, Qi Mao, Jia-Bin Huang, Yu-Ding Lu, Maneesh~Kumar
  Singh, and Ming-Hsuan Yang.
\newblock Drit++: Diverse image-to-image translation viadisentangled
  representations.
\newblock {\em International Journal of Computer Vision}, pages 1--16, 2020.

\bibitem{liu2019learning}
Xihui Liu, Guojun Yin, Jing Shao, Xiaogang Wang, and Hongsheng Li.
\newblock Learning to predict layout-to-image conditional convolutions for
  semantic image synthesis.
\newblock In {\em Advances in Neural Information Processing Systems}, 2019.

\bibitem{Logacheva_2020_ECCV}
Elizaveta Logacheva, Roman Suvorov, Oleg Khomenko, Anton Mashikhin, and Victor
  Lempitsky.
\newblock Deeplandscape: Adversarial modeling of landscape videos.
\newblock In {\em Proceedings of the European Conference on Computer Vision
  (ECCV)}, August 2020.

\bibitem{Mescheder2018ICML}
Lars Mescheder, Sebastian Nowozin, and Andreas Geiger.
\newblock Which training methods for gans do actually converge?
\newblock In {\em International Conference on Machine Learning (ICML)}, 2018.

\bibitem{Miangoleh2021Boosting}
S.~Mahdi~H. Miangoleh, Sebastian Dille, Long Mai, Sylvain Paris, and
  Ya\u{g}{\i}z Aksoy.
\newblock Boosting monocular depth estimation models to high-resolution via
  content-adaptive multi-resolution merging.
\newblock 2021.

\bibitem{Nederhood_2021_ICCV}
Cooper Nederhood, Nicholas Kolkin, Deqing Fu, and Jason Salavon.
\newblock Harnessing the conditioning sensorium for improved image translation.
\newblock In {\em Proceedings of the IEEE/CVF International Conference on
  Computer Vision (ICCV)}, pages 6752--6761, October 2021.

\bibitem{park2019SPADE}
Taesung Park, Ming-Yu Liu, Ting-Chun Wang, and Jun-Yan Zhu.
\newblock Semantic image synthesis with spatially-adaptive normalization.
\newblock In {\em Proceedings of the IEEE Conference on Computer Vision and
  Pattern Recognition}, 2019.

\bibitem{park2020swapping}
Taesung Park, Jun-Yan Zhu, Oliver Wang, Jingwan Lu, Eli Shechtman, Alexei~A.
  Efros, and Richard Zhang.
\newblock Swapping autoencoder for deep image manipulation.
\newblock In {\em Advances in Neural Information Processing Systems}, 2020.

\bibitem{pmlr-v139-ramesh21a}
Aditya Ramesh, Mikhail Pavlov, Gabriel Goh, Scott Gray, Chelsea Voss, Alec
  Radford, Mark Chen, and Ilya Sutskever.
\newblock Zero-shot text-to-image generation.
\newblock In Marina Meila and Tong Zhang, editors, {\em Proceedings of the 38th
  International Conference on Machine Learning}, volume 139 of {\em Proceedings
  of Machine Learning Research}, pages 8821--8831. PMLR, 18--24 Jul 2021.

\bibitem{Ranftl2020}
Ren\'{e} Ranftl, Katrin Lasinger, David Hafner, Konrad Schindler, and Vladlen
  Koltun.
\newblock Towards robust monocular depth estimation: Mixing datasets for
  zero-shot cross-dataset transfer.
\newblock {\em IEEE Transactions on Pattern Analysis and Machine Intelligence
  (TPAMI)}, 2020.

\bibitem{pmlr-v48-reed16}
Scott Reed, Zeynep Akata, Xinchen Yan, Lajanugen Logeswaran, Bernt Schiele, and
  Honglak Lee.
\newblock Generative adversarial text to image synthesis.
\newblock In Maria~Florina Balcan and Kilian~Q. Weinberger, editors, {\em
  Proceedings of The 33rd International Conference on Machine Learning},
  volume~48 of {\em Proceedings of Machine Learning Research}, pages
  1060--1069, New York, New York, USA, 20--22 Jun 2016. PMLR.

\bibitem{saito2020coco}
Kuniaki Saito, Kate Saenko, and Ming-Yu Liu.
\newblock Coco-funit: Few-shot unsupervised image translation with a content
  conditioned style encoder.
\newblock {\em arXiv preprint arXiv:2007.07431}, 2020.

\bibitem{schnfeld2021you}
Edgar Sch{\"o}nfeld, Vadim Sushko, Dan Zhang, Juergen Gall, Bernt Schiele, and
  Anna Khoreva.
\newblock You only need adversarial supervision for semantic image synthesis.
\newblock In {\em International Conference on Learning Representations}, 2021.

\bibitem{ming2020DFGAN}
Ming Tao, Hao Tang, Songsong Wu, Nicu Sebe, Fei Wu, Xiao-Yuan Jing, and Bingkun
  Bao.
\newblock Df-gan: Deep fusion generative adversarial networks for text-to-image
  synthesis.
\newblock {\em arXiv preprint arXiv:2008.05865}, 2020.

\bibitem{diode_dataset}
Igor Vasiljevic, Nick Kolkin, Shanyi Zhang, Ruotian Luo, Haochen Wang,
  Falcon~Z. Dai, Andrea~F. Daniele, Mohammadreza Mostajabi, Steven Basart,
  Matthew~R. Walter, and Gregory Shakhnarovich.
\newblock {DIODE}: {A} {D}ense {I}ndoor and {O}utdoor {DE}pth {D}ataset.
\newblock {\em CoRR}, abs/1908.00463, 2019.

\bibitem{wang2018pix2pixHD}
Ting-Chun Wang, Ming-Yu Liu, Jun-Yan Zhu, Andrew Tao, Jan Kautz, and Bryan
  Catanzaro.
\newblock High-resolution image synthesis and semantic manipulation with
  conditional gans.
\newblock In {\em Proceedings of the IEEE Conference on Computer Vision and
  Pattern Recognition}, 2018.

\bibitem{zhang2020resnest}
Hang Zhang, Chongruo Wu, Zhongyue Zhang, Yi Zhu, Zhi Zhang, Haibin Lin, Yue
  Sun, Tong He, Jonas Muller, R. Manmatha, Mu Li, and Alexander Smola.
\newblock Resnest: Split-attention networks.
\newblock {\em arXiv preprint}, 2020.

\bibitem{zhang2020cross}
Pan Zhang, Bo Zhang, Dong Chen, Lu Yuan, and Fang Wen.
\newblock Cross-domain correspondence learning for exemplar-based image
  translation.
\newblock In {\em Proceedings of the IEEE/CVF Conference on Computer Vision and
  Pattern Recognition}, pages 5143--5153, 2020.

\bibitem{zhang2018perceptual}
Richard Zhang, Phillip Isola, Alexei~A Efros, Eli Shechtman, and Oliver Wang.
\newblock The unreasonable effectiveness of deep features as a perceptual
  metric.
\newblock In {\em CVPR}, 2018.

\bibitem{zhu2020indomain}
Jiapeng Zhu, Yujun Shen, Deli Zhao, and Bolei Zhou.
\newblock In-domain gan inversion for real image editing.
\newblock In {\em Proceedings of European Conference on Computer Vision
  (ECCV)}, 2020.

\bibitem{zhu2017toward}
Jun-Yan Zhu, Richard Zhang, Deepak Pathak, Trevor Darrell, Alexei~A Efros,
  Oliver Wang, and Eli Shechtman.
\newblock Toward multimodal image-to-image translation.
\newblock In {\em Advances in Neural Information Processing Systems}, 2017.

\bibitem{Zhu_2020_CVPR}
Peihao Zhu, Rameen Abdal, Yipeng Qin, and Peter Wonka.
\newblock Sean: Image synthesis with semantic region-adaptive normalization.
\newblock In {\em IEEE/CVF Conference on Computer Vision and Pattern
  Recognition (CVPR)}, June 2020.

\end{thebibliography}
}

\end{document}


\title{\vspace{-3em}Supplementary materials for \\StyLandGAN: A StyleGAN based Landscape Image Synthesis using Depth-map\vspace{-3em}}  

\maketitle
\thispagestyle{empty}
\appendix

\section{Additional comparison with previous works}
\textbf{PoE-GAN.} Recently, a conditional image synthesis framework called PoE-GAN has been proposed. Through PoE-GAN, it adopts various input modalities, and even handles contradictory modalities. StyLandGAN currently uses a fixed number of modalities, and it is hard to properly reflect the results of applying contradictory conditions. Though PoE-GAN can probably be extended to incorporate depth maps, StyLandGAN has the advantage of creating contents in a progressive way. Our `2-phase inference' first translates a segmentation map, which is two-dimensional information, into a depth map, which is three-dimensional information. In each phase, our model suggests various styles while fixing previous constraints. By separating structure and style selection process within two different phases, StyLandGAN has the great advantage of allowing users to progressively embody their ideas and intentions into the landscape image to be synthesized.

\textbf{Taming Transformers.} We find that the Taming Transformers (T.T) tries a depth map input condition for image synthesis tasks. Even though T.T only tried synthesizing images with the main subject like birds and dogs, we assume this model could be extended to synthesize landscape images. However, our StyLandGAN not only focuses on using a depth map to synthesize landscape images but also suggests synthesizing various depth maps with 'two-phase inference'. Moreover, we find that our Stylegan-based architecture makes inference much faster than T.T.

\section{S2D translation result analysis}
\begin{figure*}
\begin{center}
\includegraphics[width=2\columnwidth]{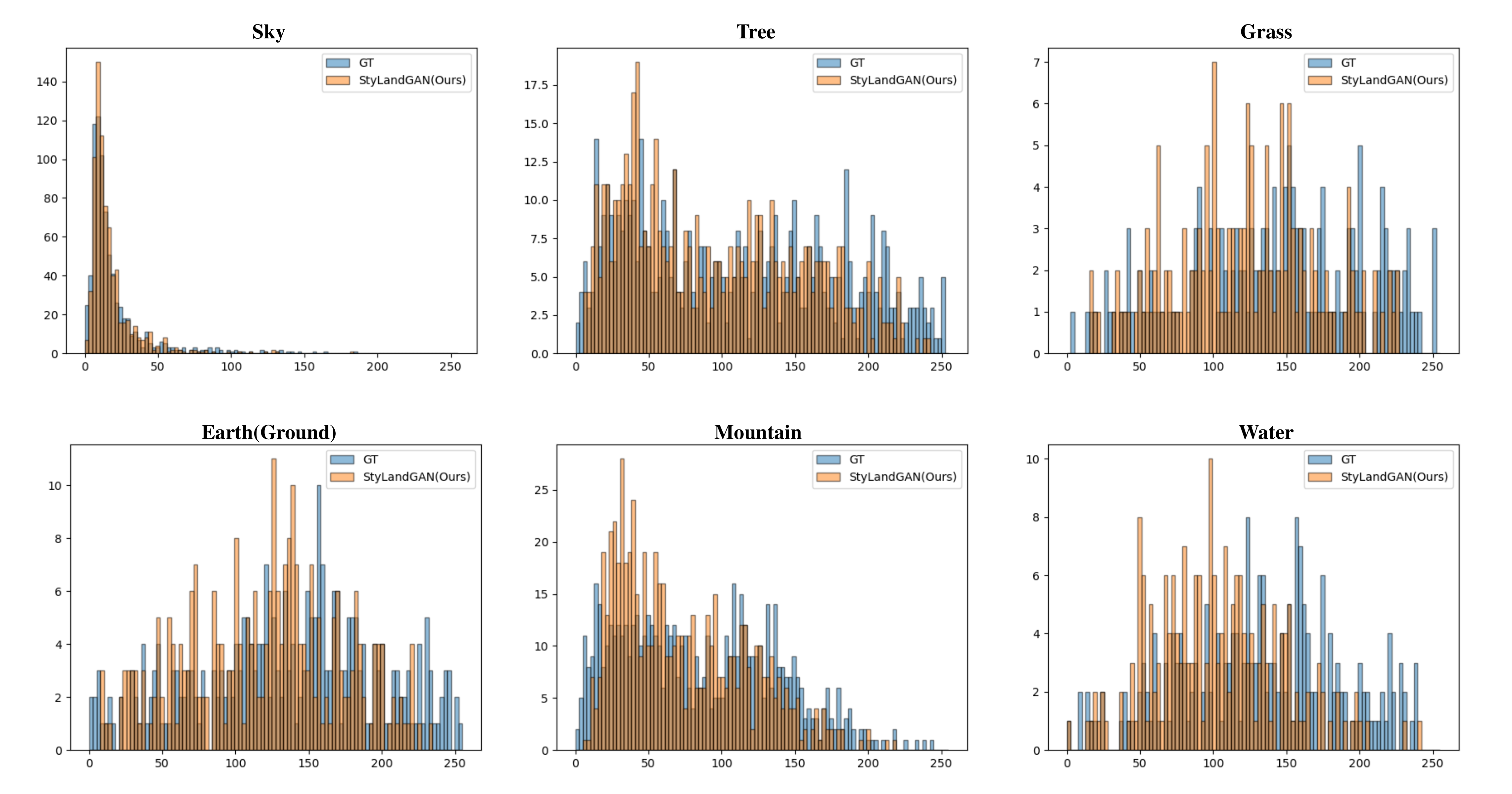}
\end{center}
   \caption{
   Distribution of mean depth maps for each representative segmentation label. We compare the depth value of the ground truth (GT) and the S2D translation result. Note that the S2D translation model synthesizes an almost similar depth distribution as GT.
   }
\label{fig:sup_depth}
\end{figure*}

Our S2D translation model should produce a plausible depth map to properly generate a final image. To verify the effectiveness of S2D translation model, we illustrate the depth distribution of the synthesized images and ground truth (GT) images. We extracted depth from Flickr-landscape test set as a ground-truth, and translated depth from semantic segmentation of test set. We divide the depth distribution of each segmentation label to observe each tendency. The distribution of the average depth value for each semantic label is shown in Fig~\ref{fig:sup_depth}. The $x$-axis represents the mean depth value of an image, where 0 represents the near area and 255 represents the far area. The $y$-axis represents the number of images. More overlaps between two distributions means that the estimated depth are similar to the ground truth, hence, realistic. Furthermore, we can also see the intuition that the sky is furthest, the mountains are closer than the sky, and the other semantics such as tree, grass, and earth lie almost everywhere in an landscape image.

\section{Additional qualitative examples}
\begin{figure*}

\begin{center}
\includegraphics[width=2\columnwidth]{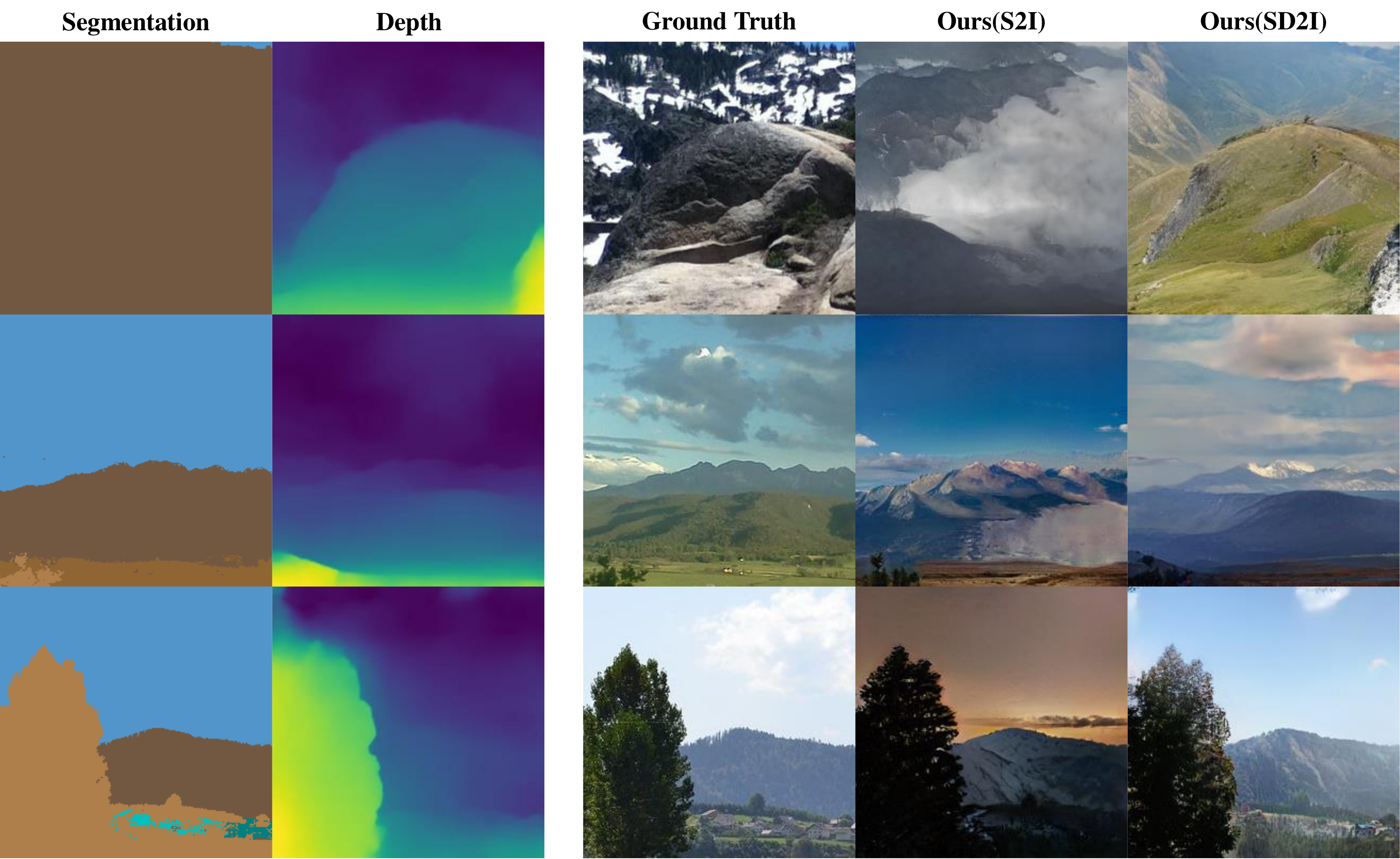}
\end{center}
   \caption{Example of generated StyLandGAN results trained on S2I and SD2I translation models. Note that even if a specific structure cannot be expressed in the segmentation map as in the first row, the depth map can have a more detailed expressions.}
\label{fig:sup_s2isd2i}
\end{figure*}

\begin{figure*}
\begin{center}
\includegraphics[width=2\columnwidth]{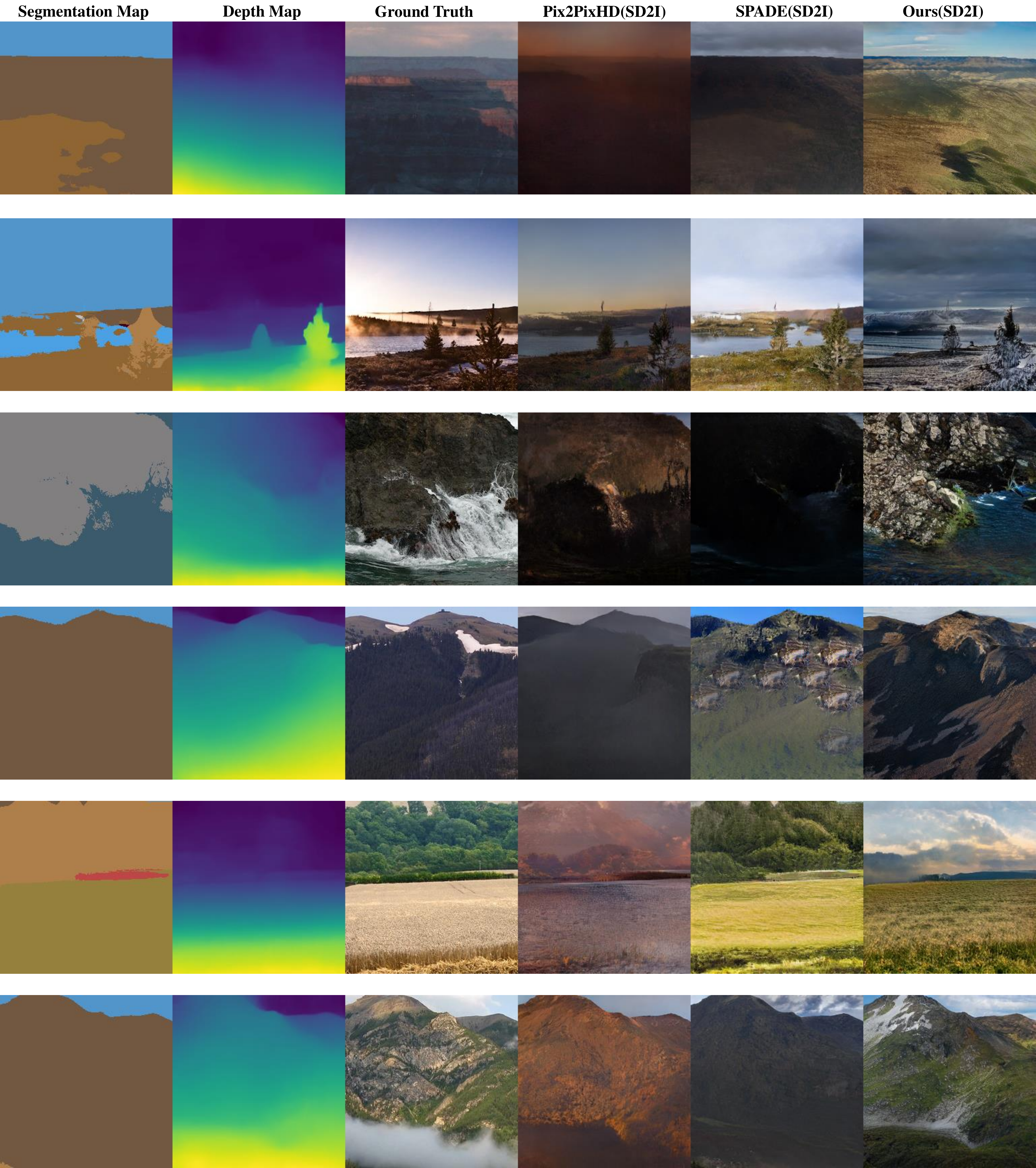}
\end{center}
   \caption{Additional qualitative comparisons of SD2I translation model on Flickr-landscape.}
\label{fig:sup_additional}
\end{figure*}

\begin{figure*}
\begin{center}
\includegraphics[width=2\columnwidth]{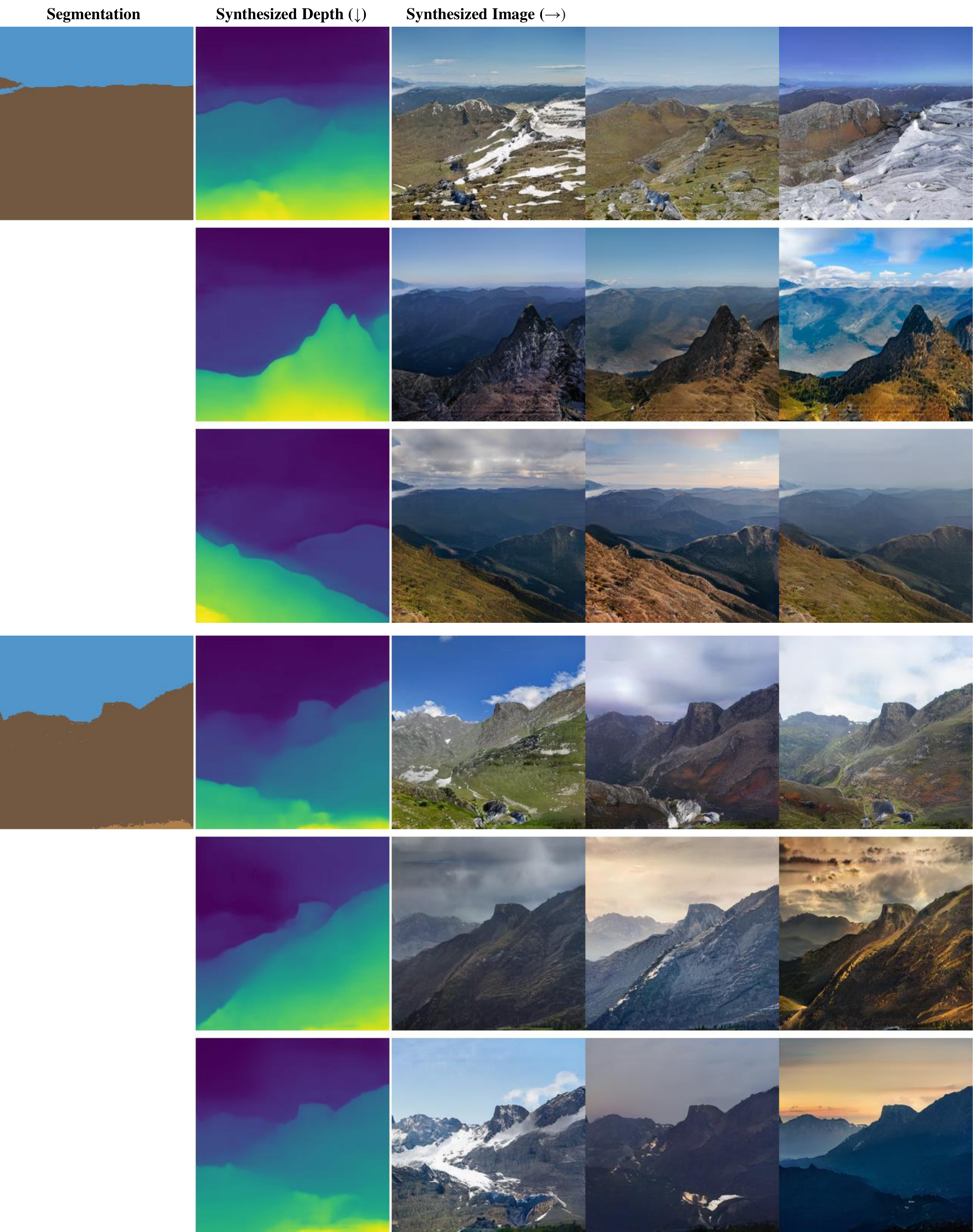}
\end{center}
   \caption{Additional results of `2-phase inference'. The S2D translation model takes the first column images and generates various depth maps as in the second column. With the first and second images, the SD2I translation model generates multimodal stylized outputs.}
\label{fig:sup_2phase}
\end{figure*}

\begin{figure*}
\begin{center}
\includegraphics[width=2\columnwidth]{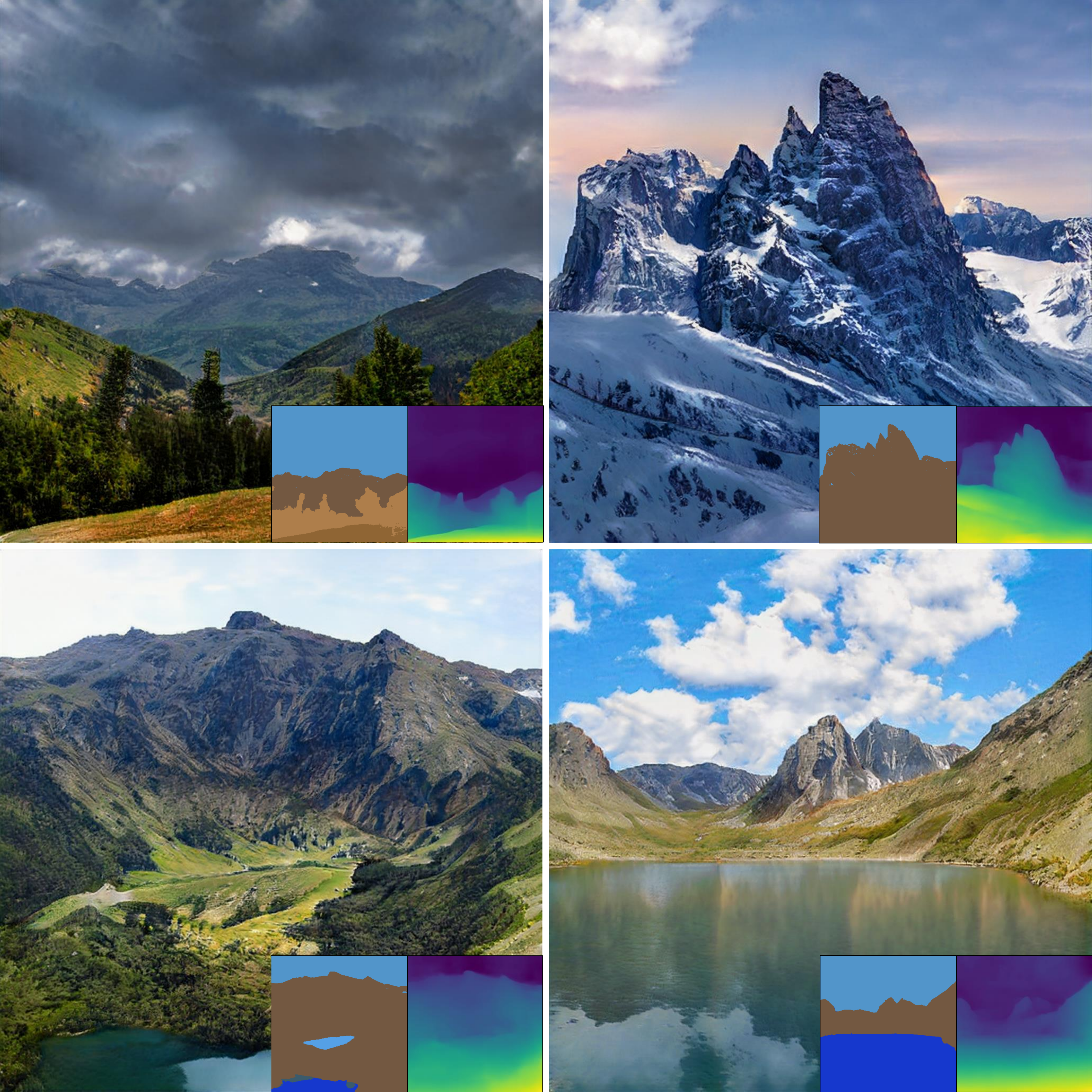}
\end{center}
   \caption{Examples of high resolution ($1024\times1024$) image synthesis with the SD2I translation model. Conditions are shown on the bottom right of the corner, which is extracted from the real landscape image.}
\label{fig:sup_highres}
\end{figure*}

We show extensive qualitative results in this section. Fig~\ref{fig:sup_s2isd2i} illustrates the advantage of a depth map by comparing S2I and SD2I translation models. Fig ~\ref{fig:sup_additional} shows the additional qualitative results of SD2I translation models with other baselines. More results on our `2-phase Inference' framework are shown in Fig~\ref{fig:sup_2phase}. Lastly, we show advantage in Fig ~\ref{fig:sup_highres}.

\section{Limitation}
\begin{figure}[h]
\begin{center}
\includegraphics[scale=0.5]{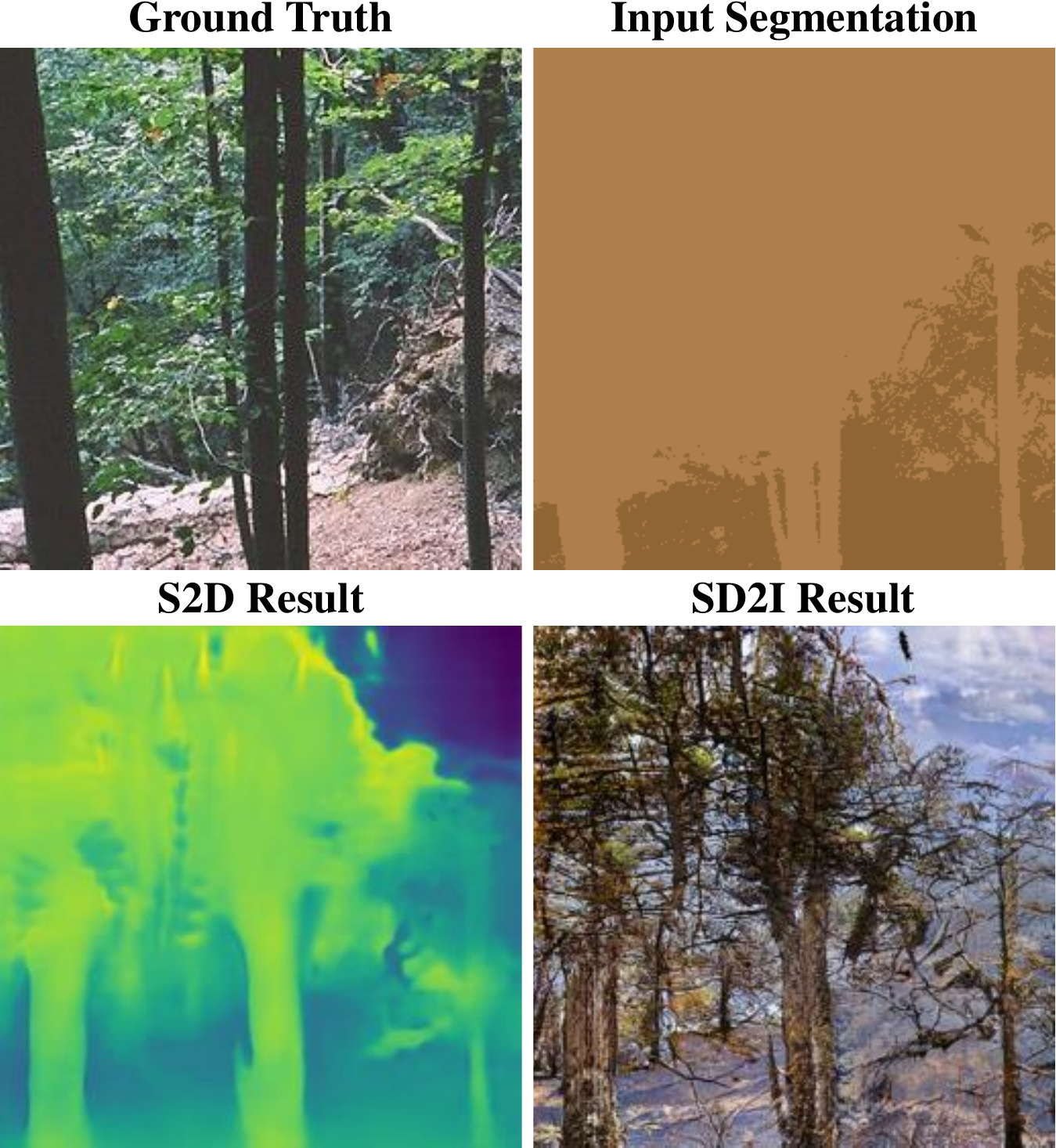}
\end{center}
   \caption{Example of our model when conditioned on close objects. Our S2D and SD2I translation models both have limitations when the object is too close to the camera.}
\label{fig:sup_limit}
\end{figure}

Although our model obtains convincing results in terms of both qualitative and quantitative evaluations compared to other models, we find out that StyLandGAN model does not work well when expressing close objects as depicted in Fig~\ref{fig:sup_limit}. Our model fails to produce plausible image due to the unrealistic depth map synthesized from segmentation of cluttered objects such as trees.

In addition, it is difficult to perfectly control the perspective of objects only by shifting the value of depth. For a more dramatic and more realistic editing of an object, its size also should be changed according to its depth value at the same time. We leave this point as future work for the development of content creation tools utilizing a depth map.


